\documentclass[11pt]{article}
\usepackage{coling2020}
\usepackage{times}
\usepackage{url}
\usepackage{latexsym}

\usepackage{float}

\usepackage{bm}
\usepackage{mathtools}
\usepackage{commath}

\usepackage{caption}

\usepackage{graphicx}
\graphicspath{ {./img/} }

\newenvironment{Figure}
  {\par\medskip\noindent\minipage{\linewidth}}
  {\endminipage\par\medskip}

\usepackage{environ}
\NewEnviron{myequation}{%
\begin{equation}
\scalebox{1.3}{$\BODY$}
\end{equation}
}

\colingfinalcopy 

\title{Morphological disambiguation from stemming data}

\author{Antoine Nzeyimana{\normalfont *} \\
  University of Massachusetts, Amherst, MA 01003, USA \\
  {\tt anthonzeyi@gmail.com} \\}

\date{}

\begin{document}
\maketitle
\begin{abstract}
  Morphological analysis and disambiguation is an important task and a crucial preprocessing step in natural language processing of morphologically rich languages. Kinyarwanda, a morphologically rich language, currently lacks tools for automated morphological analysis. While linguistically curated finite state tools can be easily developed for morphological analysis, the morphological richness of the language allows many ambiguous analyses to be produced, requiring effective disambiguation. In this paper, we propose learning to morphologically disambiguate Kinyarwanda verbal forms from a new stemming dataset collected through crowd-sourcing. Using feature engineering and a feed-forward neural network based classifier, we achieve about 89\% non-contextualized disambiguation accuracy. Our experiments reveal that inflectional properties of stems and morpheme association rules are the most discriminative features for disambiguation.
\end{abstract}

\section{Introduction}
\label{intro}

%
%
\blfootnote{
    %
    %
    %
    %
    %
    %
    \hspace{-0.65cm}  
    \textbf{*} Part of this work was done while the author was a graduate student at the University of Oregon, Eugene, OR 97403, USA \\ \\ 
    This work is licensed under a Creative Commons 
    Attribution 4.0 International License.\\
    License details: \url{http://creativecommons.org/licenses/by/4.0/}.
}

For morphologically rich languages, morphological analysis and disambiguation plays a critical role in most natural language processing (NLP) tasks. When inflections are generated by piecing together multiple morphemes, a large and sparse vocabulary is produced, requiring tools to unpack the individual morphemes for downstream NLP tasks such information extraction and machine translation. A key characteristic of these languages is that morphemes often have specific meanings (often relating to properties of the words they form or referring to contextual entities)  and their combination into words is mostly regular. Figure~\ref{analysis-figure} shows typical morphological units contained in the word 'ntuzamwibeshyeho' (\textit{Never underestimate him/her}). 

While several morphologically rich languages such as Turkish, Arabic and Modern Hebrew already have mature tools for morphological segmentation \cite{coltekin2010freely} \cite{ccoltekin2014set} \cite{itai2003corpus} \cite{habash2006magead}, Kinyarwanda still lacks appropriate tools for the task. A key limitation in the effort is the need to have high quality datasets manually annotated by language experts. With limited funding opportunities, research on NLP for low resource languages lags behind recent advancements made for NLP on high resource languages.  In this work, we leverage an easy to collect stemming dataset and transform it into a resource for morphological disambiguation. While the focus here is on Kinyarwanda verbal forms, the method can be applied to other morphologically rich languages. Collecting stemming data is much faster and less prone to errors than full morphological segmentations which require subtle linguistic knowledge. 

Through a maximum entropy model, we are able to combine morphological properties of stems with inflectional similarity information from word embeddings to accurately disambiguate candidate segmentations from a rule-based morphological analyzer. Our work here pertains to non-contextual verb-phrase disambiguation but is a key step towards contextual disambiguation. We believe that this work will allow rich morphology to be incorporated in new models for Kinyarwanda and improve downstream NLP tasks on the language.

\begin{Figure}
 \centering
 \includegraphics[scale=0.46]{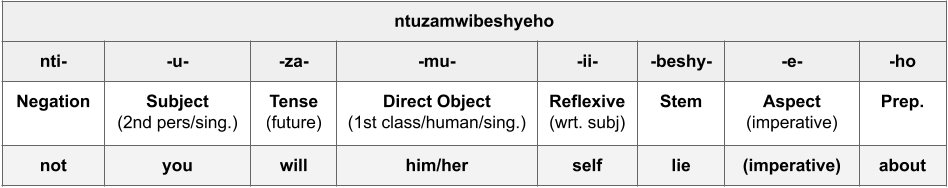}
 \captionof{figure}{\label{analysis-figure}Morphological segmentation of the word 'ntuzamwibeshyeho'. Each morpheme unit has a specific meaning or function in order to get the meaning of the whole word. The word is an inflection of the lemma 'kwibeshya' (\textit{to err}) which a derivation from 'kubeshya' (\textit{to lie}) by adding a reflexive \textit{-ii-}. Therefore, a literal morpheme-by-morpheme translation of the word would be '\textit{Never lie to yourself about him/her}' while the real meaning is '\textit{Never underestimate him/her}'.}
\end{Figure}

\begin{table}[h!]
\begin{center}
\begin{tabular}{|l|r|r|}
\hline \bf Morphological segmentation & \bf Probability \\ \hline
 - nti u/1 - za - - - mu - \textbf{ibeshy} - - - - - - - e ho & 0.184 \\
 - nti u/7 - za - - - mu - \textbf{ibeshy} - - - - - - - e ho & 0.152 \\
 - nti u/5 - za - - - mu - \textbf{ibeshy} - - - - - - - e ho & 0.137 \\
 - nti u/7 - za - - - mu ii \textbf{beshy} - - - - - y - e ho & 0.074 \\
 - nti u/1 - za - - - mu ii \textbf{beshy} - - - - - y - e ho & 0.074 \\
 - nti u/5 - za - - - mu ii \textbf{beshy} - - - - - y - e ho & 0.074 \\
 - nti u/5 - za - - - mu ii \textbf{beshy} - - - - - - - e ho & 0.062 \\
 - nti u/1 - za - - - mu ii \textbf{beshy} - - - - - - - e ho & 0.061 \\
 - nti u/7 - za - - - mu ii \textbf{beshy} - - - - - - - e ho & 0.058 \\
 - nti u/1 - za - - - mu ii \textbf{besh} - - - - - y - e ho & 0.039 \\
 - nti u/5 - za - - - mu ii \textbf{besh} - - - - - y - e ho & 0.037 \\
 - nti u/7 - za - - - mu ii \textbf{besh} - - - - - y - e ho & 0.034 \\
 - nti u/1 - za - - - mu - \textbf{ibeshy} - - - - - y - e ho & 0.009 \\
 - nti u/5 - za - - - mu - \textbf{ibeshy} - - - - - y - e ho & 0.004 \\
 - nti u/7 - za - - - mu - \textbf{ibeshy} - - - - - y - e ho & 0.001 \\ \hline
\end{tabular}
\end{center}
\caption{\label{analysis-table} Potential segmentations for the inflected verb 'ntuzamwibeshyeho' as produced by a finite state based morphological analyzer and ranked by our disambiguation method. Hyphens(-) represent potential morpheme slots that are not filled for this instance. The stem or root is shown in bold. The top three options only differ in the entity class for the subject -- \textbf{u/5} is for humans, 3rd person singular -- \textbf{u/1} also for humans but 2nd person singular -- while \textbf{u/7} is for inanimate object. Notice that the disambiguation tool predicted that the subject mostly likely is a 2nd person. The tool has also chosen to base on the derived stem '\textit{-ibeshy-}' (\textit{to err}) rather than the original root '\textit{-beshy-}' (\textit{to lie}), thus reflecting more focus on semantics. }
\end{table}

\section{Related work}

Computational morphology has been studied for decades, but most implementations and evaluations have been conducted on languages that are not related to Kinyarwanda. Finite state methods for morphological analysis have been proposed by Beesley and Karttunen \cite{karttunen2000applications} and have been popular for morphological analysis. Our morphological analyzer is based on the underlying principle of two level morphology \cite{koskenniemi1983two}, but our custom implementation does not follow the exact formalism of finite state transducers. We rather focus on refining rules that are specific to Kinyarwanda through extensive empirical examination.
In \cite{muhirwe2007computational}, a morphological alternation model for Kinyarwanda was presented using Xerox tools, but no empirical evaluation was conducted. In \cite{garrette2013real}, an experiment was conducted on learning POS taggers for Kinyarwanda and Malgasay using a small dataset of frequent words annotated by linguists. While POS tagging is an important task, morphological analysis is even more important for morphologically rich languages because it reveals more information than what can be covered by a finite set of tags. Other work on Kinyarwanda has been more linguistic in nature \cite{kimenyi1980relational}; \cite{jerro2016locative}, especially due to that Kinyarwanda is considered as a more generic prototype of the larger group of Bantu languages, owing to its rich morphology and tonal system.

The problem of morphological disambiguation has been researched on for other morphologically rich languages such as Turkish, Arabic and Modern Hebrew. Proposed methods range from statistical ones \cite{hakkani2002statistical} \cite{cohen2007joint}, to rule based approaches \cite{yuret2006learning}, and more recently using recurrent neural networks \cite{zalmout2017don}. Most of these methods are usually trained and evaluated on richly annotated tree-banks which are not available for low resource languages like Kinyarwanda. The only labeled data required by our method is a list of inflection/stem pairs which can be conveniently collected by untrained native speakers. Another major difference is also that our disambiguation focuses on uncontextualized morphology of Kinyarwanda verbal forms. We reserve contextual disambiguation and part of speech tagging for a future study.

\begin{Figure}
 \centering
 \includegraphics[scale=0.6]{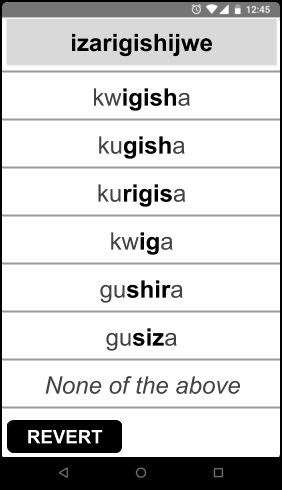}
 \captionof{figure}{\label{app-figure} A smart-phone interface is used by volunteers to stem Kinyarwanda verb forms. For instance, the analysis of the inflected form 'izarigishijwe'(\textit{...which have been made disappear}) can result in ambiguous stems -- 'igish'(kwigisha: \textit{to teach}), 'gish'(kugisha: \textit{to ask [for advice]}), 'rigis'(kurigisa: \textit{to make disappear}), 'ig'(kwiga: \textit{to study}), 'shir'(gushira: \textit{to end}) and 'siz'(gusiza: \textit{to prepare a place for construction})}
\end{Figure}

\section{Methods}

\subsection{Dataset development}

The dataset for this project comes from a crowd-sourcing effort where users labeled inflected forms of Kinyarwanda verbs with corresponding lemma. From a web-crawled corpus, our toolkit detects potential verb inflections, auto-segments them and asks volunteers to choose the right lemma from a proposed list of candidates. For convenience of use, volunteers are asked to lemmatize the inflected verbs using a simple mobile application (see Figure~\ref{app-figure}) and the labeled data are sent to a back-end server. The raw labeled dataset was filtered for potential random user inputs by using a baseline classifier and removing data for users who performed poorly on the otherwise least ambiguous instances. While more than 200 volunteers used the stemming application, only data from 37 annotators was found to be consistent and then used in this study.

\subsection{Morphological analysis}

Our morphological analysis is based on finite state methods \cite{karttunen2000applications}. Table~\ref{morphology-table} shows a repertoire of Kinyarwanda verb morphemes and examples of when they are used. The morphotactics, which dictate the ordering of morphemes, are modeled with a hierarchical graph shown in Figure~\ref{morphotactics-figure}. It is this graph of morpheme slots that make Kinyarwanda verbal system very productive. In theory, thousands of different morpheme slot sequences can be produced by this graph, but in reality, there are more semantic and syntactic restrictions.

In addition to the basic morphotactics graph model, morpho-graphemic rules and other morpheme association rules are added to the analyzer using small constraint-enforcement language. The language is expressive enough to allow a researcher to incorporate complex grammatical regularities. For example, a rule such as '\{V;NEG;ta\} $\Rightarrow$ \{!V;PRE\_IN;nti\}' prevents having two negation markers in the same verb inflection. Also, '\{V;PRE\_IN;si\} $\Rightarrow$ \{V;SUBJ;n\}' enforces the negative pre-prefix 'si' to be used only with first person singular. The rule '\{V;STEM;\#1\} $\Rightarrow$ \{V;STEM;/\textasciicircum [hzcvrz]\$/\}' limits the number of single character stems, while '\{V;STEM;/\textasciicircum gamij\$/\} $\Rightarrow$ \{V;ASP;e\}' allows the irregular verb '-gamij' (\textit{to aim}) to only take aspect marker suffix '-e'.

\begin{Figure}
 \centering
 \includegraphics[scale=0.48]{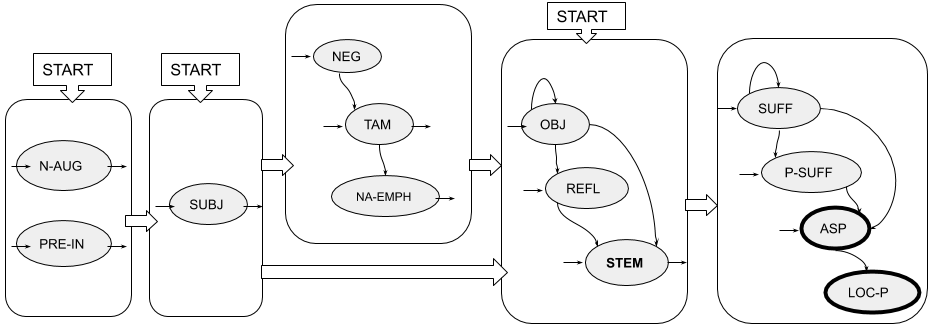}
 \captionof{figure}{\label{morphotactics-figure} A hierarchical graph model of Kinyarwanda verbal morphotactics.}
\end{Figure}

\begin{table}[h!]
\begin{center}
\begin{tabular}{|l|l|l|} 
 \hline
 \textbf{Morpheme slot} & \textbf{Morphemes} & \textbf{Example} \\ [0.5ex] 
 \hline
 1. Nominal augment (N-AUG) & \textbf{u}, a, i & \textbf{u}-a-som-ye: \\ & & uwasomye \textit{'the one who read}' \\ 
 \hline
  & si, \textbf{nti} (\textit{negative}), & \textbf{nti}-mu-a-som-ye: \\ & & ntimwasomye '\textit{you didn't read}'\\
 2. Pre-prefix (PRE-IN) & \textbf{ni} (\textit{imperative}), & \textbf{ni}-mu-som-e: nimusome '\textit{read}'\\
  & \textbf{ni} (\textit{conditional}) & \textbf{ni}-mu-som-a: nimusoma '\textit{if you read}'\\
 \hline
  & \textbf{n}, u, tu, mu, a, & \textbf{n}-a-som-ye: nasomye '\textit{I read (past)}' \\ 
 & u, \textbf{ba}, u, i, ri, a, & \textbf{ba}-za-som-a: bazasoma '\textit{they will read}'\\
 3. Subject class marker (SUBJ) & \textbf{ki}, bi, i, zi, ru, ka, & \textbf{ki}-a-som-w-ye: \\ & & cyasomwe '\textit{which was read}' \\ 
 & tu, bu, ku, ha, \textbf{ku}, & \textbf{ku}-som-a: gusoma '\textit{to read}' \\ 
 & \textbf{mi} & i-\textbf{mi}-som-ir-e: \\ & & imisomere '\textit{way of reading}' \\ 
 \hline
 4. Negation (NEG) & \textbf{ta} & tu-\textbf{ta}-som-a: tudasoma \textit{'we don't read}' \\ 
 \hline
  & i, a, ra, \textbf{ara}, za, aza, & ba-\textbf{ara}-som-ye: \\ & & barasomye \textit{'they read (past)}' \\ 
 5. Tense--Apect--Modality (TAM) & ka, raka, \textbf{kaza}, & ba-\textbf{kaza}-som-a: \\ & & bakazasoma \textit{'and they will read'} \\ 
  & ki, \textbf{racya}, oka, aka & ba-\textbf{racya}-som-a: \\ & & baracyasoma \textit{'they still read}' \\ 
 \hline
 6. Emphasizing prefix (NA-EMPH) & \textbf{na} & u-a-\textbf{na}-som-a: \\ & & wanasoma \textit{'you can \textbf{even} read}' \\ 
 \hline
 & mu, ba, wu, yi, &  \\
 7. Third object class (OBJ 3) & ri, ya, \textbf{ki}, bi, & n-ara-\textbf{ki}-ha-mu-som-ir-ye: \\
 marker (tri-transitivity) & yi, zi, ru, ka, & narakihamusomeye \\
  & tu, bu, ku, ha & \textit{'I read \textbf{it} for her while there'} \\
 \hline
 & ku, tu, ba, mu, &  \\
 8. Second object class (OBJ 2) & ba, wu, yi, ri, & n-ara-\textbf{ki}-mu-som-ir-ye: \\
 marker (di-transitivity) & ya, \textbf{ki}, bi, yi, & narakimusomeye \\
  & zi, ru, ka, tu, & \textit{'I read \textbf{it} for her'} \\
  & bu, ku, ha &  \\
 \hline
 & \textbf{n}, ku, tu, ba, &  \\
 9. First object class (OBJ 1) & mu, ba, wu, yi, & a-ara-ki-\textbf{n}-som-ir-ye: \\
 marker (transitive verbs) & ri, ya, ki, bi, & yarakinsomeye \\ 
  & yi, zi, ru, ka, & \textit{'he read it \textbf{for me}'} \\ 
  & tu, bu, ku, ha &  \\ 
 \hline
 10. Reflexive (REFL) & ii & a-\textbf{ii}-som-ir-aga:  \\ 
 & & yisomeraga \textit{'he was reading for himself'} \\ 
 \hline
 11. STEM & - & ku-\textbf{som}-a: gusoma \textit{'to read'} \\ 
 \hline
 12-13. Suffix (SUFF) & ir, ish, \textbf{ik}, iz, y & ki-ra-som-\textbf{ik}-a: \\ 
  (repeatable) & ur, uk, an & kirasomeka \textit{'it is readable}' \\ 
 \hline
 14. Passive suffix (P-SUFF) & w & ki-a-som-\textbf{w}-ye: cyasomwe \textit{'it was read by'} \\ 
 \hline
 15. Aspect marker (ASP) & a, e, \textbf{aga}, ye, i, aho & a-a-som-\textbf{aga}: yasomaga \textit{'he was reading}' \\ 
 \hline
 16. Locational post-suffix (LOC-P) & yo, ho, \textbf{mo} & a-a-som-ye-\textbf{mo}: \\ & & yasomyemo \textit{'he read inside}' \\ 
 \hline
\end{tabular}
\end{center}
\caption{\label{morphology-table} A repertoire of Kinyarwanda verbal morphemes.}
\end{table}

\subsection{Classification}

We handle morphological disambiguation as a classification problem with a variable number of classes (candidate stems) for each instance. We compute two types of features from each morphological segmentation and feed those features to a feed-forward neural network and finally produce probabilities with a softmax function. We train the network to minimize a cross-entropy loss function.

Note that, since we have a variable number of instance-specific class labels (candidate stems), the classifier is not really discriminating between a fixed set of classes but rather ranking segmentations based on the features they present. Having our labels being only the stem part of segmentation, we also need to account for the fact that there are multiple possible segmentations with the same valid stem. We account for this in our cross-entropy loss function given in Equation 1.

\begin{equation}
\mathcal{L}_{CE} = - \sum_j^M p_j log (\hat{p}_j)
\end{equation}

where:

\begin{itemize}
\item[*] M is the number of candidate segmentations produced by the rule-based analyzer
\item[*] $\hat{p}_j$ is the hypothesis probability assigned to candidate segmentation $j$ from the soft-max layer
\item[*] $p_j$ is the reference probability which we set to:
\begin{equation}
  p_j=\begin{cases}
    \frac{1}{n}, & \text{if segmentation $j$ has the right stem,}\\
     & \text{(\textit{$n$ being the number of such segmentations}).} \\
    0, & \text{otherwise}.
  \end{cases}
\end{equation}
\end{itemize}

\subsection{Feature extraction}

\subsubsection*{Similarity features}

The first type of features estimates how similar a given segmentation is to other inflections of the same stem, effectively handling the stem disambiguation part of the problem. Formally, given a candidate segmentation $x$ with a stem $s$, first we produce a set of $N$ common inflections of the same stem $\{y_i\}_i^N \in Infl(s) \cap \mathcal{V}$ by associating the stem with common standard affixes, generating the surface forms and making sure that these surface forms are part of the word embedding vocabulary $\mathcal{V}$. We choose $K$ nearest inflections to $x$ among $\{y_i\}_i^N$ (by cosine similarity) and estimate the final similarity scores as:

\begin{myequation}
f_m(\{d_e(x,y_i)\}_i^K),
\end{myequation}

where:

\begin{itemize}
\item[*] $\{\cdot\}_i^N$ notation means a set of $N$ elements indexed by $i$

\item[*] $f_m(\cdot)$ is a mean function; we use both arithmetic, geometric and harmonic means as features.

\item[*] $d_e(x,y_i)$ is the normalized angular similarity between the word embedding vectors for $x$ and $y_i$, i.e.:

\begin{myequation}
d_e(x,y_i)=\sigma(1-\frac{1}{\pi}\arccos{(\frac{\bm{e}(x)^T\bm{e}(y_i)}{\norm{\bm{e}(x)}\norm{\bm{e}(y_i)}})}),
\end{myequation} with $\bm{e(\cdot)}$ being the word embedding lookup function.

\item[*] $\sigma(\cdot)$ is a normalizing sigmoid function of the form:
\begin{myequation}
\sigma(z)=[1+exp(-8\frac{z-min_f}{Max_f-min_f})]^{-8},
\end{myequation} with $min_f$ and $Max_f$ being tunable hyper-parameters for each type of feature $f$ demarcating the active range of the feature.

\end{itemize}

Additionally, we use 2-dimensional euclidean distances of the token- and document- corpus frequencies between $x$ and $y_i$ to estimate how popular a given segmentation is in the corpus in relation to the popularity of the inflection set $\{y_i\}_i^K$:

\begin{myequation}
d_t(x,\{y_i\}_i^K)=\sigma({\frac{1}{K}\sum_{i=1}^{K}{\sqrt[]{(t_c(x)-t_c(y_i))^2+(t_d(x)-t_d(y_i))^2}}}),
\end{myequation}

where:
\begin{itemize}
\item[*] $t_c(z) = \sigma(token\_count(z))$, i.e. the sigmoid-normalized number of times $z$ appear in the corpus
\item[*] $t_d(z) = \sigma(document\_count(z))$, i.e. the sigmoid-normalized number of corpus documents containing $z$
\end{itemize}

Finally,
\begin{myequation}
f_m(\{\frac{t_c(y_i)+t_d(y_i)}{2}\}_i^K)
\end{myequation}

and

\begin{myequation}
\frac{t_c(x)+t_d(x)}{2}
\end{myequation}

are included as separate features.

\subsubsection*{Morphological indicator features}

The second type of features evaluates the appropriateness of "morphological features" present in a given segmentation versus typical features associated with its stem in the training dataset. These indicator features include the use of special morphemes, morpheme associations and special morpho-graphemic rules inherent in the segmentation. For example, passivization (transformation from active to passive form) is expressed by a special suffix but not all verbs can be used in passive form. The same applies to transitivity (the number and type of object pronouns a verb can take), the use of special suffixes, personal pronouns, locatives, and so on. Essentially, the $M$ linguistically-motivated indicator features $f_i(x)$ are compared to their selection ratio scores in the training dataset. By selection ratio scores, we mean:

\begin{myequation}
f_m(\{\sigma(\frac{chosen[f_i,s]}{proposed[f_i,s]})\}_i^M)
\end{myequation}

and separately

\begin{myequation}
\sigma(\frac{chosen[f_i,s]}{proposed[f_i,s]}),
\end{myequation}

where:

\begin{itemize}

\item[*] $chosen[f_i,s]$ is the number of times stem $s$ has been chosen as the valid stem for any morphological segmentation having morphological feature $f_i$.

\item[*] $proposed[f_i,s]$ is the number of times stem $s$ has been proposed (either chosen or rejected) among candidate lemmas for any segmentation having morphological feature $f_i$.

The list of indicator features $f_i$ used in our experiments are given in Appendix A.

\end{itemize}

\subsubsection*{Feature extraction example}

Here we present a working example of how features are extracted.
Given the input the input word '\textbf{gatwikirwa}' to be lemmatized by the annotator, we follow the following steps to extract the training data.

\textbf{Step 1. Morphological analysis:} The morphological analyzer first produces candidate segmentations as provided in Table 2. The morphological indicator features (from Appendix A) of each candidate analysis are shown in the same table.
\begin{table}[H]
\begin{center}
\begin{tabular}{|c | l | l|} 
 \hline
 \textbf{No} & \textbf{Morphological analysis (morpheme sequence)} & \textbf{Indicator features from Appendix A} \\ [0.5ex] 
 \hline
1 & - - ka - - - - - tu - ik - - ir - - - w a - & 1, 5, 17, 24, 26 \\
\hline
2 & - - ka - - - - - tu ii kir - - - - - - w a - & 1, 5, 8, 24 \\
\hline
3 & - - ka - - - - - - - twik - - ir - - - w a - & 1, 17, 24, 26 \\
 \hline
4 & - - ka - - - - - - - twikirw - - - - - y - a - & 1, 23 \\
 \hline
5 & - - ka - - - - - - - twikirw - - - - - - - a - & 1 \\
 \hline
6 & - - ka - - - - - - - twikir - - - - - - w a - & 1, 24 \\
 \hline
\end{tabular}
\end{center}
\caption{\label{analyses-table} Example of morphological analysis for input word '\textbf{gatwikirwa}' (\textit{1. covered or 2. burned} )}
\end{table}

\textbf{Step 2. Annotator input:} The annotator is presented with a list of candidate lemma to choose from:
\begin{table}[H]
\begin{center}
\begin{tabular}{|l|} 
 \hline
 \textbf{gatwikirwa} \\
 \hline
1. kw\textbf{ik}a (\textit{to sink, fall, drop)} \\ 
2. gu\textbf{kir}a (\textit{1. to be healed, releaved or 2. to be rich)} \\ 
3. gu\textbf{twik}a (\textit{to burn)} \\ 
4. gu\textbf{twikirw}a (\textit{to be covered)} \\ 
5. gu\textbf{twikir}a (\textit{to cover)} \\ 
 \hline
\end{tabular}
\end{center}
\end{table}
Assuming the annotator chooses 'gu\textbf{twikir}a' as the correct lemma, then the annotated raw data consists of the 5 labeled pairs: gatwikirwa/\textbf{ik}:0/1, gatwikirwa/\textbf{kir}:0/1, gatwikirwa/\textbf{twik}:0/1, gatwikirwa/\textbf{twikirw}:0/2 and gatwikirwa/\textbf{twikir}:1/1.

\textbf{Step 3. Generate inflection sets $S_j=\{y_i\}_i^N$ for each candidate stem $j$:} The morphological analyser formulates a set of common inflections for each candidate stem:
\begin{table}[H]
\begin{center}
\begin{tabular}{|l|} 
 \hline
1. \textbf{ik}: $S_1=\{$kwika, turitse, turika, twitse, twikaga, yaritse, arika, ...$\}$ \\ 
2. \textbf{kir}: $S_2=\{$gukira, arakize, turakira, dukize, bakize, tuzakire, ...$\}$ \\ 
3. \textbf{twik}: $S_3=\{$gutwika, uzatwika, uzatwike, hatwitsemo, atwikamo, ...$\}$ \\ 
4. \textbf{twikirw}: $S_4=\{$gutwikirwa, tugatwikirwa, yaratwikiwe, agatwikirwa, ...$\}$ \\ 
5. \textbf{twikir}: $S_5=\{$gutwikira, yatwikiriza, yatwikirije, akitwikira, ...$\}$ \\ 
 \hline
\end{tabular}
\end{center}
\end{table}

\textbf{Step 4. Compute similarity features:} The input word $x=$'gatwikirwa' is then paired with each of the nearest $K$ entries $y_i$ in the inflection sets $S_j$ to compute similarity scores $d_e(x,y_i)$, $t_c(.)$, $t_d(.)$ and $d_t(x,S_j)$ according to equations 4-8. Example values are provided below.
\begin{table}[H]
\begin{center}
\begin{tabular}{|l|l|} 
 \hline
$d_e($'gatwikirwa', 'kwika'$) = 0.075$ & $d_t($'gatwikirwa', 'kwika'$) = 0.05$ \\ 
$d_e($'gatwikirwa', 'gukira'$) = 0.003$ & $d_t($'gatwikirwa', 'gukira'$) = 0.01$ \\ 
$d_e($'gatwikirwa', 'gutwika'$) = 0.822$ & $d_t($'gatwikirwa', 'gutwika'$) = 0.61$ \\ 
$d_e($'gatwikirwa', 'gutwikirwa'$) = 0.992$ & $d_t($'gatwikirwa', 'gutwikirwa'$) = 0.81$ \\ 
$d_e($'gatwikirwa', 'gutwikira'$) = 0.986$ & $d_t($'gatwikirwa', 'gutwikira'$) = 0.72$ \\ 
 \hline
\end{tabular}
\end{center}
\end{table}

\textbf{Step 5. Compute morphological indicator features:} The morphological indicator features shown in Table 3 will be used to populate the 'proposed' and 'chosen' tables in equations 9-10. For instance, given the morphological analyzer output in Table 3 and annotator choice as \textbf{gatwikirwa/twikir}, the tables are going to be populated as in the following example:
\begin{table}[H]
\begin{center}
\begin{tabular}{|l|l|l|} 
 \hline
proposed[24, 'twikir'] & incremented by 1 & \\
 \hline
chosen[24, 'twikir'] & incremented by 1 & i.e. '\textit{cover}' is more likely to be passivized \\
 \hline
proposed[8, 'kir'] & incremented by 1 & \\
 \hline
chosen[8, 'kir'] & decremented by 1 & i.e. '\textit{be rich}' is less likely to be subject reflexive \\
 \hline
proposed[1, 'twikirw'] & incremented by 2 & \\
 \hline
chosen[1, 'twikirw'] & decremented by 2 & i.e. the passivized form is just demoted \\
 \hline
\end{tabular}
\end{center}
\end{table}

\textbf{Step 6. Neural network input/output for training:} For every candidate segmentation, the neural network receives an input of 64 real values composed of 3 from equation 3, 1 from equation 6, 3 from equation 7, 1 from equation 8, 3 means (geometric, arithmetic and harmonic) of all similarity features, 3 values from equation 9, 47 values from equation 10 (1 for each indicator feature in Appendix A) and 2 binary (0 or 1) features indicating the popularity of the lemma in two lexical resources (small/restricted and large). An extra sigmoid-normalized weighted average similar to equation 6 was added to account for the popularity of the type of inflection used to generate each element in the inflection set.
In this example, the target probability $p_j=1$ for the pair gatwikirwa/\textbf{twikir} and $p_j=0$ for the other 4 pairs.

\section{Experimental setup}

The first step in our experiments was to generate stem morphological indicator features from user annotations as explained in section 3.4. The second step involves preparing the dataset for training and evaluation. After features are extracted, we split the data between in training and validation set and then train a baseline classifier using only the data from user annotations. We up-sample by 4 factors the annotations from the best trained annotator who is equipped with subtle linguistic understanding of Kinyarwanda verb morphology. We use the baseline classifier to then predict the stem for the entire unlabeled vocabulary of Kinyarwanda verbal forms. We rank the predicted stems by prediction uncertainty (entropy). The most uncertain instances are sent back to annotators for labeling in a batched active learning fashion \cite{settles2009active}. For active learning, we send batches of the top 10000 uncertain samples for which the entropy $H > 1$. We also enrich our labeled set with the most confident predictions in a semi-supervised manner. For semi-supervised learning, we only take examples for which the baseline model has labeled with at least 0.95 top probability ($P_1$), having at least 3 competing stems, $(P_1 - P_2) > 0.95$ and entropy $H < 0.1$.

After expanding our labeled data through active learning and semi-supervised learning, we repeat the first step of feature extraction to form our final training and evaluation dataset, which contains about 170,000 examples. We split our dataset into training, development and test set in the ratio of 70\%+15\%+15\% respectively. All our models are trained with gradient descent using ADAM update rule \cite{kingma2014adam} using 0.01 learning rate in 256-sized mini-batches and for 50
 epochs. For our best fine-tuned model, we re-train the model with large batches of 4000 examples each using LAMB method \cite{you2019large} for 100 more epochs. Since all features are precomputed, training each feed-forward neural network takes less than 5 minutes on an quad-core machine. We use POSIX C and C++11 for this project with the only external dependency being Eigen matrix algebra library \footnote{\url{http://eigen.tuxfamily.org/index.php?title=Main_Page}}.

\section{Experimental results and discussion}

\textbf{Model size} -- We evaluated the model robustness by varying the number of hidden units in the feed-forward neural nets (Table~\ref{models-table}). Surprisingly, the size of the of the model doesn't affect the performance. Even a small network of two hidden layers of 6 and 3 units respectively achieves almost the same accuracy as the network of 3 layers of 32,16,8 hidden units. We also observed very little over-fitting, having the same level of accuracy on both training, development and test set. We believe that this persistent performance is probably due to the semi-supervised method we used and possibly that the summarizing features (i.e. $f_m(\cdot)$) precomputed explain most of the label variations.

\begin{table}[h!]
\begin{center}
\begin{tabular}{|r | c | c|} 
 \hline
 \textbf{Size of feedforward layers} & \textbf{Morphological Embedding} & \textbf{FastText Embedding} \\ [0.5ex] 
 \hline
 64-32-16-8 & 89.32 & 89.30 \\ 
 64-32-8 & 89.39 & 89.30 \\ 
 64-6-3 & 89.28 & 89.19 \\ [1ex] 
 \hline
\end{tabular}
\end{center}
\caption{\label{models-table} Test set accuracy for various model sizes using all 64 input features}
\end{table}

\textbf{Feature subsets} -- We then evaluated different feature subsets to assess which ones had greater impact on the final performance. All the results presented in Table~\ref{features-table} used a the small model of two layer, 64-6-3 feed-forward units. The three statistics (arithmetic, geometric and harmonic means) of morphological features account for most of the accuracy while using the individual morphological features under-performs. This is probably due to that the small nature of neural network used doesn't allow it to effectively learn these statistics. The difference in the performance of inflectional similarity features may be attributed to the differences in the two pre-trained word embeddings used. The vocabulary of our "Morpho" embeddings is almost as twice as big than the fastText \cite{bojanowski2017enriching} one, even though they are trained on the same corpus and both are based on the Skip-Gram model \cite{mikolov2013distributed}. So, comparing them requires carefully setting proper hyper-parameters $min_f$ and $Max_f$ for the normalizing function $\sigma(\cdot)$ in equation 4.

\begin{table}[h!]
\begin{center}
\begin{tabular}{|l | c | c|} 
 \hline
 \textbf{Feature subset} & \textbf{Morpho. Embedding} & \textbf{FastText Embedding} \\ [0.5ex] 
 \hline
 Summary statistics & 87.92 & 88.29 \\ 
 Inflectional  similarity with word embeddings & 66.47 & 68.74 \\ 
 Inflectional similarity with token frequency & 51.94 & 44.24 \\ 
 Morphological features summary & \underline{88.16} & \underline{88.36} \\ 
 Morphological feature details & 85.52 & 85.30 \\ [1ex] 
 \hline
 All features combined and training longer & \textbf{89.51} & \textbf{89.55} \\ [1ex] 
 \hline
\end{tabular}
\end{center}
\caption{\label{features-table} Test set accuracy using various feature subsets using 6-3 hidden units. The fine-tuned model was trained longer using 32-8 MLP setup}
\end{table}

\begin{table}[h!]
\begin{center}
\begin{tabular}{|l | c | c|} 
 \hline
 \textbf{Annotator training level} & \textbf{Uniformly random samples} & \textbf{Most uncertainty samples} \\ [0.5ex] 
 \hline
 Graduate student & 84.9 & 61.9 \\ 
 College graduate & 80.7 & - \\ 
 High school graduate & 77.9 & 55.2 \\ [1ex] 
 \hline
\end{tabular}
\end{center}
\caption{\label{annotators-table} Relative performance (training set accuracy) of different annotators with varying levels of linguistic training.}
\end{table}

\textbf{Annotator performance} -- Our final evaluation looked at how our fine-tune model rated different individual annotator labels depending on their linguistic training level and the mode of active learning used (Table~\ref{annotators-table}). The 3 reported annotators in the table were identified contacts of the author and together contributed more than 30\% of the labeled data. Our interpretation of the results is that the model might be relying too much on easy examples pulled in through semi-supervised learning and noise introduced by individual annotators.  The level of annotator training also has a clear impact on the performance.

\textbf{Sources of ambiguity} -- There are inherently multiple sources of ambiguity when one encounters a Kinyarwanda verbal expression. Achieving full disambiguation requires having access to complete contextual information. This information may even be encoded only in the tonal system \cite{kimenyi2002tonal} and thus unavailable in written form. In fact, reading written Kinyarwanda requires careful real-time disambiguation by the reader because tones are not marked in text. Contextual information is also needed for semantic disambiguation. For example, the verb \textbf{'yarigishije'} can mean both 'a-ara-igish-iz-ye' (\textit{he taught}) or 'a-a-rigis-iz-ye' (\textit{he made disappear}). Without the semantic context, both segmentations are possible. Sentence level disambiguation may also benefit from contextual agreements through the Bantu noun class system. Our annotation process is also affected by lemmatization ambiguity and the blurred boundary between inflection and derivation. For example it is subjective whether the verbs \textbf{kwivuga} 'ku-ii-vug-a' (\textit{to talk about self}), \textbf{kuvuza} 'ku-vug-y-a' (\textit{'to make sound with (some object)'}) and \textbf{kuvugisha} 'ku-vug-ish-a' (\textit{to talk to (someone)}) are themselves lemma forms or just inflections of \textbf{kuvuga} 'ku-vug-a' (\textit{to talk}).

\section{Conclusion}

This work focused on morphological disambiguation of Kinyarwanda verb forms using maximum entropy model on new crowd-sourced stemming dataset. High disambiguation accuracy was achieved through careful feature engineering. Intuitively curated inflectional features emerged as important parsimonious predictors. Future work should look at how to directly use morpheme embedding methods as a way to more generically represent both semantics and morphology in a unified form. Achieving total disambiguation ultimately requires complete contextual information which may not be available in written form. 

\section*{Acknowledgements}

This research was partly made possible by access to the S3C Laboratory facility in the Department of Geography at the University of Oregon. We thank the volunteers who contributed to the stemming dataset through Kinyarwanda Stemmer mobile application. We also thank the anonymous reviewers for their insightful feedback.

\raggedbottom

\bibliographystyle{coling}
\bibliography{coling2020}

\section*{Appendix A. Morphological indicator features used by the classifier}

\begin{table}[H]
\begin{center}
\begin{tabular}{|c|l|l|} 
 \hline
 \textbf{No} & \textbf{Feature key} & \textbf{Feature explanation} \\ [0.5ex] 
 \hline
1 & f.with\_subjects & Has a subject marker \\
\hline
2 & f.missing\_subj & Doesn’t have a subject marker \\
\hline
3 & f.with\_human\_subjects & Has human subject marker \\
\hline
4 & f.with\_location\_subjects & Has locative subject marker \\
\hline
5 & f.with\_objects & Has object marker \\
\hline
6 & f.with\_human\_objects & Has human object marker \\
\hline
7 & f.with\_location\_objects & Has locative object marker \\
\hline
8 & f.trans\_any\_refl & Has reflexive marker \\
\hline
9 & f.trans\_any\_obj3 & Has any object marker (transitive) \\
\hline
10 & f.trans\_any\_obj23 & Has two object markers (ditransivive) \\
\hline
11 & f.trans\_any\_obj123 & Has three object markers (tritransitive) \\
\hline
12 & f.obj2\_obj3\_suff & Ditransitive with any suffix \\
\hline
13 & f.obj3\_suff & Transitive with any suffix \\
\hline
14 & f.obj2\_obj3\_refl\_suff & Ditransitive with reflexive marker and any suffix \\
\hline
15 & f.obj3\_refl\_suff & Transitive with reflexive marker and any suffix \\
\hline
16 & f.suff\_ish & Has suffix \textbf{-ish} \\
\hline
17 & f.suff\_ir & Has suffix \textbf{-ir} \\
\hline
18 & f.suff\_iz & Has suffix \textbf{-iz} \\
\hline
19 & f.suff\_an & Has suffix \textbf{-an} \\
\hline
20 & f.suff\_ik & Has suffix \textbf{-ik} \\
\hline
21 & f.suff\_uk & Has suffix \textbf{-uk} \\
\hline
22 & f.suff\_ur & Has suffix \textbf{-ur} \\
\hline
23 & f.suff\_y & Has suffix \textbf{-y} \\
\hline
24 & f.suff\_w & Has passive suffix \textbf{-w} \\
\hline
25 & f.suff\_ir\_y & Has suffixes \textbf{-ir} followed by \textbf{-y} \\
\hline
26 & f.suff\_ir\_w & Has suffixes \textbf{-ir} followed by \textbf{-w} \\
\hline
27 & f.suff\_an\_y & Has suffixes \textbf{-an} followed by \textbf{-y} \\
\hline
28 & f.suff\_iz\_y & Has suffixes \textbf{-iz} followed by \textbf{-y} \\
\hline
29 & f.suff\_y\_w & Has suffixes \textbf{-y} followed by \textbf{-w} \\
\hline
30 & f.post\_suff & Has locative post-suffix \\
\hline
31 & f.mg\_rule\_r\_y\_none & Uses suffix rule \textbf{r + y} $\rightarrow$ \textbf{y} e.g. biragoye: bi-ra-gor-ye (\textit{it is difficult}) \\
\hline
32 & f.mg\_rule\_r\_y\_z & Uses suffix rule \textbf{r + y} $\rightarrow$ \textbf{z} e.g. yarabuze: a-ara-bur-ye (\textit{went missing}) \\
\hline
33 & f.obj3\_ka\_ku & Has object markers \textbf{-ka} or \textbf{-ku} \\
\hline
34 & f.suff\_ish\_ish & Has suffixes \textbf{-ish} followed by \textbf{-ish} \\
\hline
35 & f.suff\_ir\_ir & Has suffixes \textbf{-ir} followed by \textbf{-ir} \\
\hline
36 & f.comb\_obj3\_suff\_w & Has an object marker, a suffix and a passive suffix \textbf{-w} \\
\hline
37 & f.with\_nloc\_obj\_no\_suf & Has non-locative object marker but without a suffix \\
\hline
38 & f.suff1\_suff2 & Has two suffixes \\
\hline
39 & f.suff1\_suff2\_suff3 & Has three suffixes \\
\hline
40 & f.ni\_imperative & Uses the imperative prefix \textbf{ni-} \\
\hline
41 & f.ni\_conditional & Uses the conditional prefix \textbf{ni-} \\
\hline
42 & f.mg\_rule\_t\_y\_s & Uses suffixation rule \textbf{t + y} $\rightarrow$ \textbf{s} e.g. narose: n-a-rot-ye (\textit{I dreamed}) \\
\hline
43 & f.mg\_rule\_t\_y\_sh & Uses suffixation rule \textbf{t + y} $\rightarrow$ \textbf{sh} e.g. nafashe: n-a-fat-ye (\textit{I took}) \\
\hline
44 & f.suff\_an\_ir & Has suffixes \textbf{-an} followed by \textbf{-ir} \\
\hline
45 & f.suff\_ur\_y & Has suffixes \textbf{-ur} followed by \textbf{-y} \\
\hline
46 & f.suff\_ur\_w & Has suffixes \textbf{-ur} followed by \textbf{-w} \\
\hline
47 & f.suff\_uk\_y & Has suffixes \textbf{-uk} followed by \textbf{-y} \\
 \hline
\end{tabular}
\end{center}
\end{table}

\end{document}